\documentclass[conference]{IEEEtran}
\IEEEoverridecommandlockouts
\usepackage{cite}
\usepackage{amsmath,amssymb,amsfonts}
\usepackage{graphicx}
\usepackage{textcomp}
\def\BibTeX{{\rm B\kern-.05em{\sc i\kern-.025em b}\kern-.08em
    T\kern-.1667em\lower.7ex\hbox{E}\kern-.125emX}}

\usepackage[caption=false,font=footnotesize]{subfig}

\usepackage{bm}
\usepackage{hyperref}
\usepackage[nolist]{acronym}
\usepackage{comment}

\usepackage[capitalise]{cleveref}
\usepackage{algorithm}
\usepackage[usenames,dvipsnames]{xcolor}
\usepackage{algorithmic}
\usepackage{ulem}
\hypersetup{
    colorlinks=true,
    linkcolor={red!70!green},  
    citecolor={blue!50!green}, 
    urlcolor={blue!80!black}   
}
\floatname{algorithm}{Algorithm}

\usepackage{fancyhdr}
\begin{document}


\title{Ensemble Bayesian Decision Making with Redundant Deep Perceptual Control Policies
\thanks{This work was supported by Amazon Web Services (AWS) and Komatsu Ltd.}
}

\author{
\IEEEauthorblockN{Keuntaek Lee, Ziyi Wang, Bogdan Vlahov, Harleen Brar, and Evangelos A. Theodorou}
\IEEEauthorblockA{\textit{Autonomous Control and Decision Systems Laboratory} \\
\textit{Georgia Institute of Technology}\\
Atlanta, GA 30332, USA \\
keuntaek.lee@gatech.edu}
}

\maketitle
\thispagestyle{fancy}
\pagestyle{empty}

\begin{acronym}
\acro{NN}{Neural Network}
\acro{DL}{Deep Learning}
\acro{dGPS}{differential GPS}
\acro{hl}{\textit{heteroscedastic} loss}
\acro{KL}{Kullback-Leibler}
\acro{IL}{Imitation Learning}
\acro{iLQG/MPC-DDP}{iterative Linear Quadratic Gaussian/Model Predictive Control Differential Dynamic Programming}
\acro{MPC}{Model Predictive Control}
\acro{DDP}{Differential Dynamic Programming}
\acro{BNN}{Bayesian Neural Network}
\end{acronym}

\begin{abstract}
This work presents a novel ensemble of \acp{BNN} for control of safety-critical systems. Decision making for safety-critical systems is challenging due to performance requirements with significant consequences in the event of failure. In practice, failure of such systems can be avoided by introducing redundancies of control. \acp{NN} are generally not used for safety-critical systems as they can behave in unexpected ways in response to novel inputs. In addition, there may not be any indication as to when they will fail. \acp{BNN} have been recognized for their ability to produce not only viable outputs but also provide a measure of uncertainty in these outputs.  This work combines the knowledge of prediction uncertainty obtained from \acp{BNN} and ensemble control for a redundant control methodology. Our technique is applied to an agile autonomous driving task. Multiple \acp{BNN} are trained to control a vehicle in an end-to-end fashion on different sensor inputs provided by the system. We show that an individual network is successful in maneuvering around the track but crashes in the presence of unforeseen input noise. Our proposed ensemble of \acp{BNN} shows successful task performance even in the event of multiple sensor failures.

Supplementary video: \href{https://youtu.be/poRbH__kB2M}{https://youtu.be/poRbH$\_\_$kB2M}
\end{abstract}


\section{Introduction}
\label{sec:intro}

\acresetall{}

\acp{NN} are currently one of the most powerful tools for solving difficult problems in decision making such as playing Go \cite{silver2016mastering} or medical diagnosis \cite{medical,medical2}. Notably, \acp{NN} are able to perform rapid and complex computations through relatively simple nonlinear calculations and massive parallel structures. Because of this, \acp{NN} have been applied to a variety of difficult classification and regression problems from object detection to robotics. One such task that benefits from the performance of \acp{NN} is end-to-end imitation learning for autonomous driving. In this task, difficulty arises in mapping sensor inputs into driving commands \cite{PanRSS18}. Previous work, such as \cite{PanRSS18,LevineVisuomotor,bojarski2016end}, shows the successful use of end-to-end imitation learning with applications to autonomous driving and manipulation with visual inputs. Under the imitation learning settings, a system can efficiently learn a task, guided by an expert. However, much of the previous work does not investigate the robustness of the learned end-to-end model to compromised sensors.

Although \acp{NN} are capable of successfully completing difficult tasks in a variety of applications, they are not without drawbacks. One drawback is that it can be nearly impossible to determine what the output of the \ac{NN} will be given an input without using the \ac{NN} itself.  This is due to the nonlinear computational structure and a large number of parameters. Another drawback is that \acp{NN} are also heavily reliant on data; they generally can not make use of prior knowledge such as dynamics models. When confronted with a new input, the output of the network can vary drastically, even if there are similarities to inputs from training data. Even small perturbations to the input can alter the output of Deep \acp{NN} \cite{onepixelattack,adversarialattacks} and the Deep \acp{NN} are easily fooled \cite{nguyen2015deep}. This means we do not have a consistent mapping from inputs to outputs. In the context of safety, traditional \acp{NN} do not provide a measure of uncertainty of the output. 

In recent years, however, new improvements have been made on probabilistic \acp{NN}. \acp{BNN} are a probabilistic network structure that produces a distribution of outputs rather than a single output. This output distribution provides valuable information showing how certain or uncertain the output is. With the ability to measure uncertainty, \acp{NN} become a viable option for ensemble techniques used for decision making. Ensemble techniques consist of a set of hypotheses from which they choose one as the output. In \cite{EnsembleCIO}, the ensembles of perturbed models are used to perform robust trajectory optimization with respect to model uncertainty. The work in \cite{Lakshminarayanannips17} demonstrated that a simple ensemble model can effectively approximate the predictive uncertainty of \ac{DL} if the objective function obeys a proper scoring rule. This method used multiple \acp{NN} with different initializations to serve as individual models of an ensemble for approximating predictive uncertainty. However, the obtained predictive uncertainty was not directly used for improving the performance of the target task. 

\begin{figure}[b]
    \centering
    \subfloat{\includegraphics[width=0.22\textwidth]{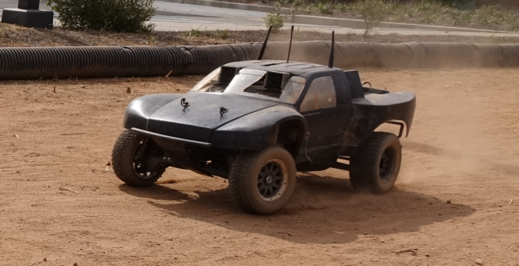}}
    \hfil
    \subfloat{\includegraphics[width=0.255\textwidth]{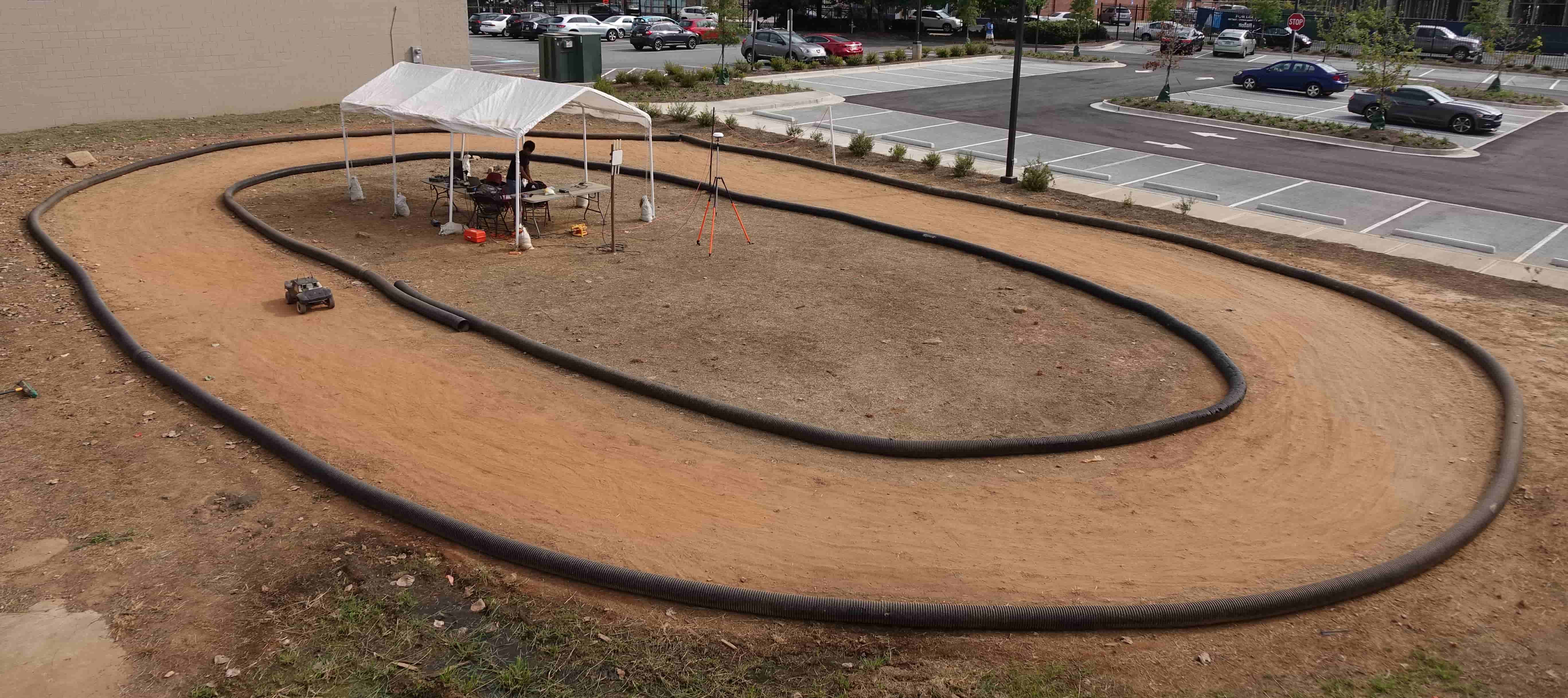}}
    \caption{\textit{Left:} The $1/5$ scaled ground vehicle for autonomous driving and racing. \textit{Right:} The oval track used for experiments.}
    \label{fig:car}
\end{figure}

With knowledge of the uncertainty of each hypothesis, ensemble techniques can be used in safety-critical systems where the failure of the system causes tragic results. In this work, we propose a novel ensemble of end-to-end \acp{BNN} to provide an elegant solution to sensor failure in safety-critical systems. Our method is applied to the platform seen in \cref{fig:car}, with the task of agile autonomous driving.  With aggressive maneuvers on harsh terrain, sensors can fail from damage or are unable to operate effectively with rapidly changing conditions.

The rest of this paper is organized as follows: in \cref{sec:background}, we provide background information for the key ideas used in this paper. In \cref{sec:ensemble_bayesian} we introduce our ensemble \acp{BNN} structures and provide the algorithm for decision making. We discuss the expert used for data collection in \cref{sec:data_collect} and present results in \cref{sec:experiments}. Finally, we give our conclusions and discuss future work in \cref{sec:conclusion}.

\section{Background}
\label{sec:background}
In this section, we will cover a few concepts central to our proposed research solution. In order to detect sensor failure, we will be measuring uncertainty in \ac{NN} models, in comparison to the more traditional approach of sensor fault detection before passing sensor information to a \ac{NN}. There exists an extensive literature on sensor failure detection \cite{imufailure, isermann1984process, qin2009data} that demonstrates its application in various fields. However, this approach requires knowledge of the expected sensor outputs to determine whether a reading is normal or faulty. Our approach lets the learned model itself address this problem by utilizing the probabilistic counterpart of the traditional \acp{NN}, namely \acp{BNN}. This Bayesian approach of deep learning removes the need for any beforehand knowledge of expected sensor outputs. A brief overview of types of uncertainty is given to provide the motivation of \acp{BNN}. We finish by briefly covering \acl{IL}, which is the method we use to train our models.

\subsection{Aleatoric and Epistemic Uncertainty}
Model uncertainty can be classified into two major categories \cite{WhatUncertainties}: aleatoric and epistemic uncertainty. Aleatoric uncertainty is a result of the model's inability to fully describe the environment, while epistemic uncertainty is a result of the inability to acquire unlimited data. In the first case, uncertainty arises when different outcomes are obtained even with the same experimental setup. The source of this type of uncertainty is the hidden variables that can not be perfectly characterized or measured. Epistemic uncertainty arises when the model is presented data not seen previously. The source for this type of uncertainty is a data set that does not fully cover the sample space. In application, it is not possible to completely eliminate either form of uncertainty as we do not have access to a perfect model or unlimited data.

The origin of aleatoric uncertainty suggests that we should be able to train a model to output this type of uncertainty given data. Meanwhile, we should also be able to measure the epistemic uncertainty of a model through some form of sampling. In this paper, the total predictive uncertainty is calculated to be the combination of both uncertainty types.

\subsection{\aclp{BNN}}
\label{subsec:bnn}
Currently, there exist two popular methods to obtain a predictive probability distribution in the deep learning literature. The first technique uses Bayesian Backpropagation \cite{pmlr-v37-blundell15}, which assigns a probability prior, usually Gaussian, to the weights in the network. The network is trained by minimizing the \ac{KL} divergence between the distribution on the weights and the true Bayesian posterior distribution.

The second approach uses dropout layers to produce a predictive distribution resulting from a probabilistic network structure \cite{pmlr-v48-gal16}. The Monte Carlo dropout approach is adopted in this paper since the alternative approach requires at least doubling the number of parameters in the network, which makes it difficult to run a large scale convolutional \aclp{NN} with only the computational resources on-board the vehicle. Using the existing \ac{NN} structure with dropout added to every weight layer, weights in the network are randomly dropped with a certain probability. At every forward pass, we sample a dropout mask from a Bernoulli distribution to determine weights dropped in each layer. During the backward pass, only the remaining weights are updated. The outputs of the network then become a Gaussian Distribution, returning the mean and variance of the prediction values. When trained with the loss function described in \cref{subsec:ensemble_structure} and \cref{subsec:implementation}, these outputs becomes a combination of aleatoric and epistemic uncertainty. The work in \cite{pmlr-v48-gal16} shows the mathematical equivalence between an approximated deep Gaussian process and a \ac{NN} with arbitrary depth and nonlinearities when dropout layers are applied before and after every weight layer. The output distribution is estimated with Monte Carlo sampling, which can be done in parallel to reduce run-time.

\subsection{\acl{IL}}
\label{subsec:imitation_learning}
\ac{IL}, also called ``learning from demonstration'', is a type of supervised machine learning. \ac{IL} is often used when the optimal solution to the task is not easily accessible or too computationally expensive to run in real time. \ac{IL} algorithms assume that an oracle policy or expert is available. The expert can utilize resources that are unavailable for the imitation learner at test time, such as additional sensory information and computing power. In the case of autonomous driving, the expert can be a sophisticated optimal control algorithm or an experienced human driver. The observation-action or state-action pairs generated by the expert is then used to train the imitation learner. The goal of \ac{IL} is to mimic the expert's behavior as well as possible. In \cite{PanRSS18} \ac{IL}'s ability to perform the autonomous driving task with low-cost sensors is demonstrated on real-world experiments.

In the traditional formulation, the goal is often to find a policy $\pi : \mathcal{O} \rightarrow \mathcal{U}$ that minimizes an expected loss or maximizes an expected reward over a discrete finite time horizon $H$:
\begin{equation}
    \label{eq:min_cost}
    \min_{\pi} \mathbb{E}_{p_{\pi}} \left[ \sum_{t=0}^{H-1} l(x_t,u_t)\right],
\end{equation}
where $x_t \in \mathcal{S}, u_t \in \mathcal{U}$, and $\mathcal{S,O,U}$ are the state, observation and admissible control spaces respectively. $l$ is the immediate loss function. $p_{\pi}$ is the joint distribution of $x_t,u_t$, and $o_t \in \mathcal{O}$ for the policy $\pi$ for $t = 0, ..., H - 1$.

For imitation learning, this equation changes slightly. The goal now becomes to learn a policy that minimizes a loss function that characterizes the difference between the learned model and the expert policy $\pi^{*}$ rather than the most optimal $\pi$:

\begin{equation}
    \label{eq:min_cost_imitation}
    \pi_{NN} = \arg\min_{\pi} \mathbb{E}_{p_{\pi^{*}}} \left[ l(x_t,u_t)\right],
\end{equation}
where $u_t = \pi(o_t)$ and $\pi_{NN}$ denotes the neural network policy chosen.
In our work, we trained our networks with batch imitation learning.

\section{Ensemble Bayesian Decision Making}
\label{sec:ensemble_bayesian}
\subsection{Problem Formulation}
\label{subsec:problem_formulation}
The main problem considered in this paper is the autonomous driving task for a $1/5$ scaled ground vehicle (\cref{fig:car}) using deep neural network-based end-to-end control policies under the sensor failure cases. As mentioned in \cref{sec:intro}, many applications of deep end-to-end control do not provide a principle solution to sensor failure. Most self-driving cars today depend on different kinds of sensors including Lidar, Radar, GPS, and cameras. However, in the real world, these sensors are vulnerable to noise. In one example, \ac{dGPS} is widely used for autonomous driving to obtain global positions in the world frame. Despite many advances in GPS technology, there is always the probability that the GPS signal jumps or slightly diverges from the true position. In areas with obstacles such as tall buildings or indoor parking lots, GPS tends to fail altogether. Additionally, cameras are sensitive to light conditions and interference from external sources. In autonomous driving, even a slight shift of the GPS data or an obscured camera may cause the car to pass the center line of the road with significant consequences. To avoid these failures, system redundancy is crucial for the safe operation of autonomous driving.

System redundancy is commonly applied in many safety-critical applications, where multiple backup systems exist to prevent catastrophic failure from one faulty component, as shown in \cite{stein2003respect}. Redundancy is usually achieved by either duplicating the same system or using different systems that perform the same task. It is easy to have duplicative systems available in case of failure, but they are vulnerable to faults of the underlying system. Dissimilar backups, where different hardware, software, and control laws are used in each backup system, can alleviate this problem, but it is hard to determine how the backup systems are prioritized when a failure occurs. 

In this work, an ensemble Bayesian decision making process is used to provide system redundancy. Multiple \acp{BNN} are implemented on the vehicle, each taking in a different sensory input and having the capability of performing the task on its own. Each \ac{BNN} is trained end-to-end, learning the low-level control actions from each sensor input. When one or more of the sensors is compromised, the associated predictive uncertainty to the failed sensor should see a significant increase that causes the system to switch to the remaining functional networks.

\subsection{Ensemble Structure}
\label{subsec:ensemble_structure}

\begin{figure}[h]
    \centering
    \includegraphics[width=\linewidth]{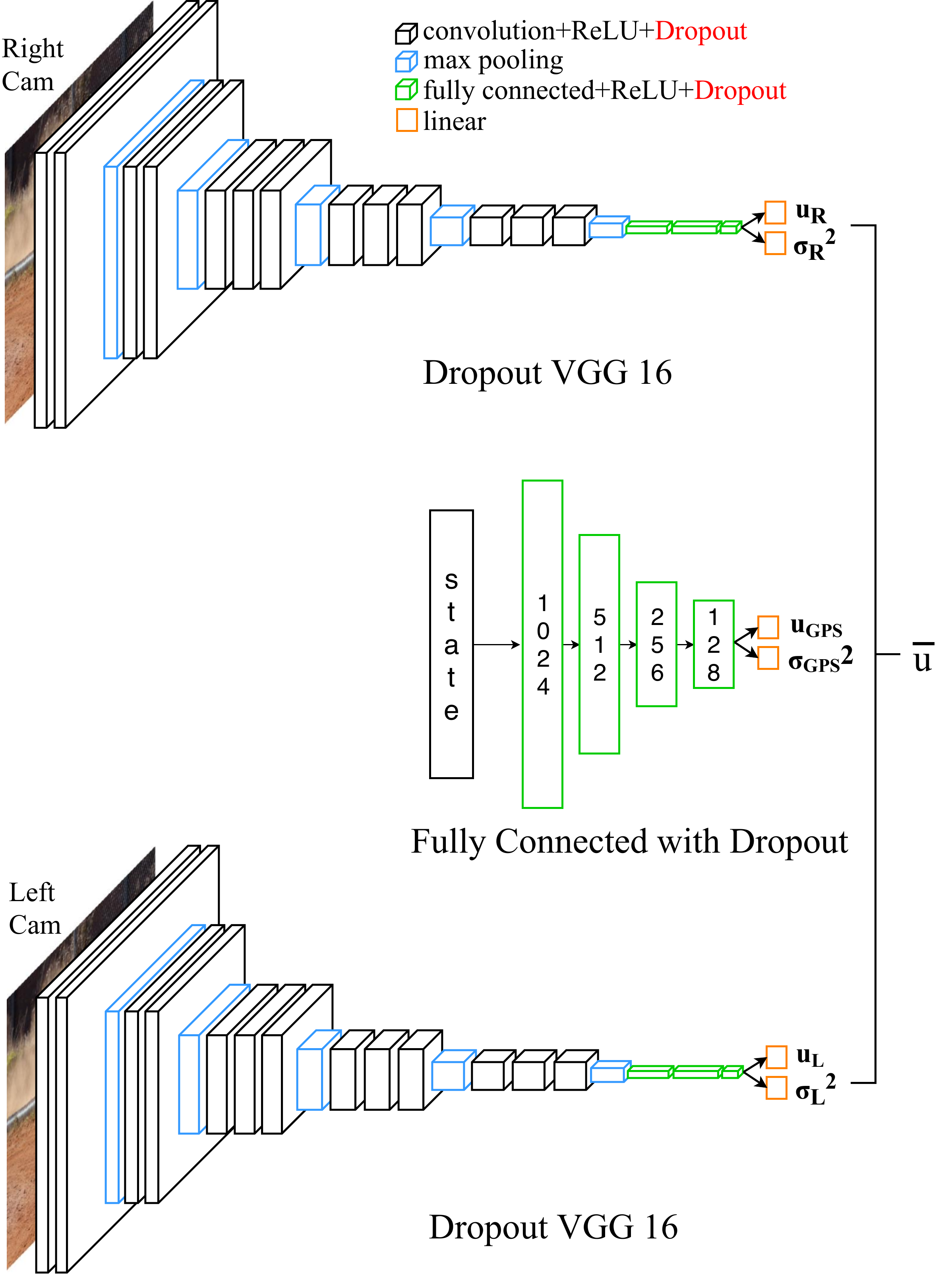}
    \caption{Ensemble Network Structure composed of end-to-end Bayesian Networks.}
    \label{fig:network}
\end{figure}

\begin{figure}[b]
    \centering
    \subfloat{\includegraphics[width=0.23\textwidth]{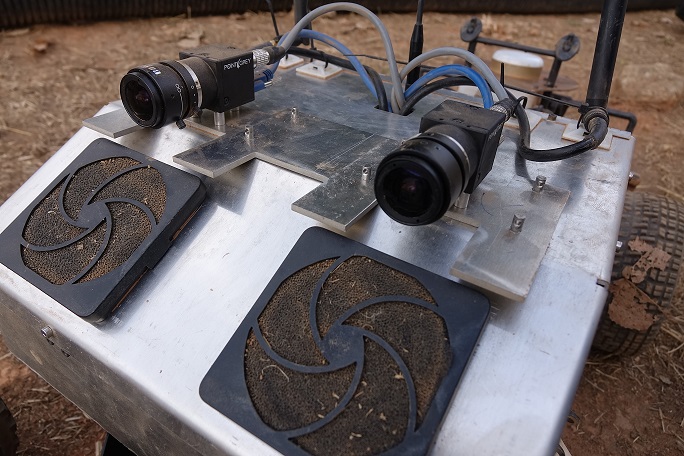}}
    \label{fig:camera}
    \subfloat{\includegraphics[width=0.23\textwidth]{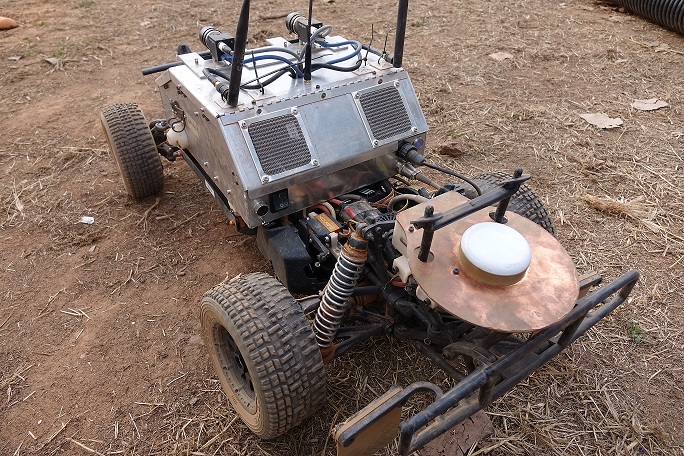}}
    \label{fig:img_gps_system}
    \caption{Platform sensors connected to the on-board computer. \textit{Left}: Two vision cameras, 1280x1024, 70fps, global shutter, synchronously triggered. \textit{Right}: RTK-corrected Hemisphere Eclipse P307 GPS module, position at 20Hz.
}\label{fig:sensors}
\end{figure}

Our ensemble consists of 3 \acp{BNN}. Each \ac{BNN} differs in their network input as well as their network structure.  They output the mean, $\hat{u}$, and the variance, $\hat{\sigma}^2$, of the model caused by the aleatoric uncertainty. The first \ac{BNN} is trained on the fully-observable state data gathered from GPS module. Its network structure is fully connected with ReLU activation functions and layers of width 1024, 512, 256, and 128, respectively. The second network is trained on images taken from the camera on the left side of the vehicle shown in \cref{fig:sensors}.
It is using the VGG 16 \cite{VGG}-like network, with modifications to include dropout at each layer as well as output the variance, $\hat{\sigma}^2$ stemming from aleatoric uncertainty \cite{lee2019earlyfailure}. This deep neural network is composed of $\sim$30 million trainable parameters, depending on the size of the input. The last network is trained on images taken from the camera on the right side of the vehicle shown in \cref{fig:sensors} and has the same network structure as the second network. 
The overall network structure can be seen in \cref{fig:network}.

To ensure that the outputs of each \ac{BNN} are the mean and variance due to aleatoric uncertainty, the \acl{hl} function is used. This loss function used is defined in \cite{WhatUncertainties}, and is as follows: 
\begin{equation}
    \label{eq:h_loss}
    \mathcal{L}(\pi) = \frac{1}{2\hat{\sigma}^2} ||u^{*} - \hat{u} ||^2 + \frac{1}{2}\log(\hat{\sigma}^2),
\end{equation}
where $\pi$ is the current policy of the network, $u^{*}$ is the expert's action for input $x$, $\hat{u}$ is the aleatoric mean for input $x$, $\hat{\sigma}^2$ is the aleatoric variance for input $x$. To see how $\hat{\sigma}^2$ is a measure of the aleatoric variance, let us think about how aleatoric variance should behave. We would like that the predictions ($\hat{u}$) that are close to the expert's output ($u^{*}$) -- resulting in a low residual error -- have low aleatoric variance.  Predictions that are far away from the expert's output -- resulting in a high residual error -- should also have high aleatoric variance. To minimize \cref{eq:h_loss} when the residual error is high, $\hat{\sigma}^2$ must increase so that the residual error does not have a strong impact on the loss. When the residual error is small, it is observed that $\hat{\sigma}^2$ also needs to be small in order to minimize \cref{eq:h_loss}. Intuitively, since $\hat{\sigma}^2$ follows the increase or decrease of the residual error to obtain a minimal loss, \cite{WhatUncertainties} concludes that $\hat{\sigma}^2$ is at least an approximation of the aleatoric variance. In practice, the \acl{hl} function is modified slightly to:
\begin{equation}
    \label{eq:h_loss_exp}
    \mathcal{L}(\pi) = \frac{1}{2} \exp(-s) ||u^{*} - \hat{u} ||^2 + \frac{1}{2}s,
\end{equation}
where $s = \log(\hat{\sigma}^2)$. By regressing with $s$, we avoid a potential 'division by $0$' error and can still easily calculate $\hat{\sigma}^2$. These means and variances are then used to find the output of the ensemble network as described in the next section. 

\subsection{Implementation} \label{subsec:implementation}
To get a better calibrated uncertainty measure, we used Concrete Dropout \cite{Gal2017Concrete}, which allows for automatic tuning of the dropout probability in large models.
The output of the redundant system structure is calculated in \cref{alg:ensemble}. As described in \cref{subsec:bnn}, we need to sample our networks multiple times in order to generate the output predictive distribution. In \cref{alg:ensemble}, instead of conducting multiple runs of the network $i$ on a single input $x_i$, we duplicate each input $n_{MC}$ times to create an input sequence $\mathbf{X}_{i}$, and then input this sequence through the network, where $n_{MC}$ is the number of samples used for Monte Carlo sampling.
The two outputs of network $i$ are a vector of control commands $\mathbf{\hat{u}_i}$ and a vector of the aleatoric variances $\bm{\hat{\sigma}_i}^2$. Using these vector outputs the overall variance (aleatoric and epistemic combined) of each \ac{BNN} is calculated with the following equation (step 4 in \cref{alg:ensemble}):
\begin{equation}
    \label{eq:variance}
    \sigma_i^2 = Var(u_i) \approx \underbrace{\frac{1}{K} \sum_{k=1}^{K}\hat{u}_{i_k}^2 - \left(\frac{1}{K} \sum_{k=1}^{K}\hat{u}_{i_k} \right)^2}_{epistemic} + \underbrace{\frac{1}{K} \sum_{k=1}^{K}\hat{\sigma}_{i_k}^2}_{aleatoric} ,
\end{equation}
where $u_i$ is the output of network $i$, $\hat{u}_i$ is the aleatoric mean of network $i$, $\hat{\sigma}_i^2$ is the aleatoric variance  of network $i$, and ($\mathbf{\hat{u}_{i}}, \bm{\hat{\sigma}_{i}^2}$) is the set of $K$ sampled outputs from network~$i$. The control of each network $i$ is calculated as the mean of that network's sampled outputs:
\begin{equation}
    \label{eq:output}
    u_i = \frac{1}{K} \sum_{k=1}^{K}\hat{u}_{{i_k}}.
\end{equation}

Note that, in step 3 in \cref{alg:ensemble}, the computation (prediction) of all Bayesian Networks happens in parallel.

Finally the output $\bar{u}$ is chosen from the network $i$ with the lowest variance, $\sigma_i^2$, as shown in step 5.

There are two possible ways to do the ensemble Bayesian decision making. One approach is weighting individual network policies with some weights that inversely proportional to their variance. However, this weighting approach is not an optimal solution when the system encounters a multi-modal situation. For example, if one of the network policies tries to drive a vehicle to the left and the other tries to steer it to the right and they have almost equal weights, the ensemble of the policies will guide the vehicle to go straight. This can lead to a tragic result if the network policies made a prediction to drive either left or right and there is an obstacle on the straight.

The other way to do ensemble Bayesian decision making is to pick the best decision according to its confidence, as we proposed in step 5 in \cref{alg:ensemble}. This approach helps avoid choosing non-optimal control policies in multi-modal decision cases.

\begin{algorithm} [t]
\caption{Ensemble \aclp{BNN}}
\begin{algorithmic}[1]
\label{alg:ensemble}
\REQUIRE $\newline \mathbf{x}_{L} \text{: Image from left camera; }\newline \mathbf{x}_{R} \text{: Image from right camera;}\newline \mathbf{x}_{GPS} \text{: States from GPS}$
\WHILE {\textit{Task not failed}}
\STATE $\mathbf{X}_{L}, \mathbf{X}_{R}, \mathbf{X}_{GPS} \leftarrow \text{Duplicate }\mathbf{x}_{L}, \mathbf{x}_{R}, \mathbf{x}_{GPS}$
\STATE $\mathbf{\hat{u}}_{L}, \bm{\hat{\sigma}}^2_{L} \leftarrow \text{DropoutVGG16(}\mathbf{X}_{L}\text{)}$
\\ $\mathbf{\hat{u}}_{R}, \bm{\hat{\sigma}}^2_{R} \leftarrow \text{DropoutVGG16(}\mathbf{X}_{R}\text{)}$
\\ $\mathbf{\hat{u}}_{GPS}, \bm{\hat{\sigma}}^2_{GPS} \leftarrow \text{FC4(}\mathbf{X}_{GPS}\text{)}$
\STATE $u_L, u_R, u_{GPS} \leftarrow \text{Mean(} \mathbf{\hat{u}}_L, \mathbf{\hat{u}}_R, \mathbf{\hat{u}}_{GPS}\text{)}$
\\ $\sigma^2_L, \sigma^2_R, \sigma^2_{GPS} \leftarrow \text{Var(}\mathbf{\hat{u}}_L, \mathbf{\hat{u}}_R, \mathbf{\hat{u}}_{GPS}, \bm{\hat{\sigma}}^2_{L}, \bm{\hat{\sigma}}^2_{R}, \bm{\hat{\sigma}}^2_{GPS}\text{)}$
\STATE $\bar{u} \leftarrow \text{MinVar(}u_L, u_R, u_{GPS}, \sigma^2_L, \sigma^2_R, \sigma^2_{GPS} \text{)}$
\ENDWHILE
\ENSURE $\bar{u}\text{: Steering command for the vehicle}$
\end{algorithmic}
\end{algorithm}

\section{Data Collection}
\label{sec:data_collect}

In order to train each learner (\ac{BNN}) in the ensemble, the \ac{iLQG/MPC-DDP} \cite{MPCDDP} algorithm was used as an expert. \ac{DDP} is an algorithm that uses second-order approximations of the cost function and system dynamics around a nominal trajectory to solve the Bellman equation. The optimal control policy is then used to update the nominal trajectory. Running \ac{DDP} in \ac{MPC} fashion means that at every timestep, only the first control action is executed by the system, and the control policy is re-optimized at the next timestep when new state information is received. For our long-term autonomous driving task, we used the receding horizon DDP \cite{MPCDDP}.

Using GPS data, the expert had the following state space [$p_x$, $p_y$, $\theta$, $\psi$, $V_x$, $V_y$, $\dot{\theta}$] as input, where $p_x$ and $p_y$ are global positions in the world frame, $\theta$ and $\psi$ are the heading and roll angle of the car, $V_x$ and $V_y$ are the body frame longitudinal and lateral velocities, and $\dot{\theta}$ is the derivative of the heading angle.

We considered the cost function for the optimal controller composed of an arbitrary state-dependent cost and a quadratic control cost. The state-dependent term was designed to stay in the center of the track ($p_{x,des}$, $p_{y,des}$) while maintaining the desired forward velocity $V_{x,des}$. We set $V_{x,des}$ as 5m/s when we collect data. For the control cost, we used the same weights for both throttle and steering.

The expert drove around an oval track seen in \cref{fig:car} for 100 laps in one direction to gather data for each learner. As it drove around, the GPS data and truncated 64x128x3 RGB images from the left and right cameras were saved in order to train each of the learner models described in \cref{sec:ensemble_bayesian}. For training of the Dropout VGG 16 Net \cite{lee2019earlyfailure}, we did not use any data augmentation technique (random flips, rotations, contrast, brightness, saturation, jitter, etc.) but we truncated and cropped the original 4k image to reduce the size of it to 64x128x3. All of our models were trained in batch with Tensorflow \footnote{\href{https://www.tensorflow.org/}{https://www.tensorflow.org/}} using the Adam optimizer \cite{adam} and the heteroscedastic loss in \cref{eq:h_loss_exp}.

\section{Experiments/Results}
\label{sec:experiments}
\begin{figure}[t]
    \centering
    \subfloat[Raw image\label{fig:img_clean}]{\includegraphics[width=0.23\textwidth]{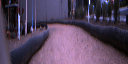}}
    \vspace{1mm}
    \subfloat[Image with artificial noise \label{fig:img_noise}]{\includegraphics[width=0.23\textwidth]{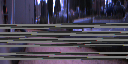}}
    \caption{Artificial noise injected to an input image at test time.}\label{fig:noisy_img}
\end{figure}

All computation was executed on-board the vehicle with our NVIDIA GeForce GTX 1050 GPU and we were able to obtain 10 Monte Carlo samples ($n_{MC}=10$) of the ensemble in real time (20Hz).
We injected artificial noise signal to each sensor similar to a real-world situation in which a sensor malfunctions. The position noise was sampled from a uniform distribution to make the "new position" appear to be outside of the track. This is commonly seen as GPS data jumps from one location to another. For images, rows of gray bars were added to simulate periodic noise caused by electro-mechanical interference during the image capturing process (\cref{fig:img_noise}).

First, we tested each \ac{BNN} in the ensemble network without any artificial noise injected. Each \ac{BNN} was able to drive the vehicle autonomously until the vehicle's batteries run out and there were no failures. In all experiments, we considered the failure cases as when the vehicle crashes to the boundaries of the track and cannot move forward.

\begin{figure}[hb]
    \centering
    \subfloat[Expert MPC-DDP\label{fig:mpcddp}]{\includegraphics[width=0.24\textwidth]{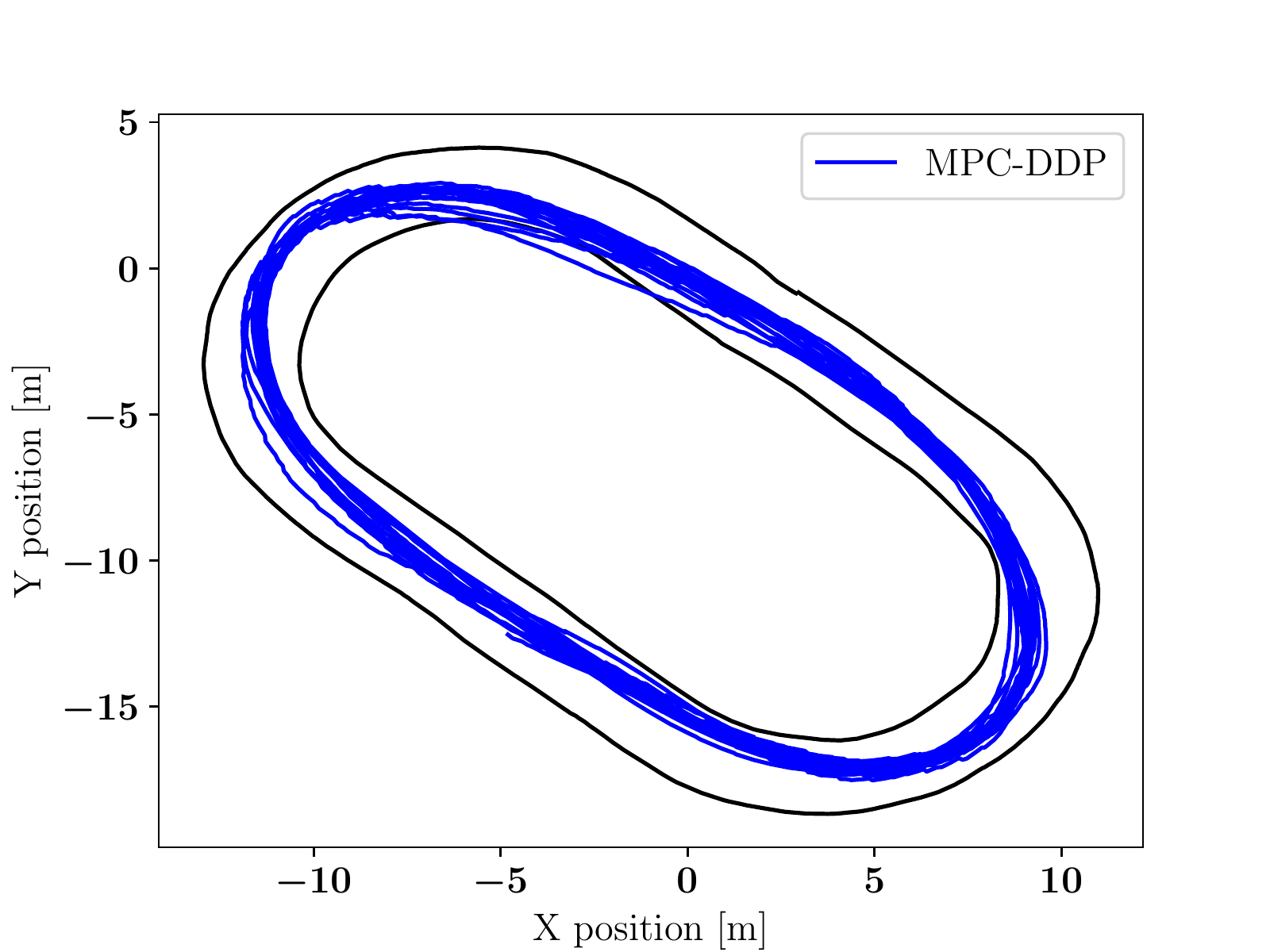}}
    \subfloat[Left camera \ac{NN} \label{fig:img_left}]{\includegraphics[width=0.24\textwidth]{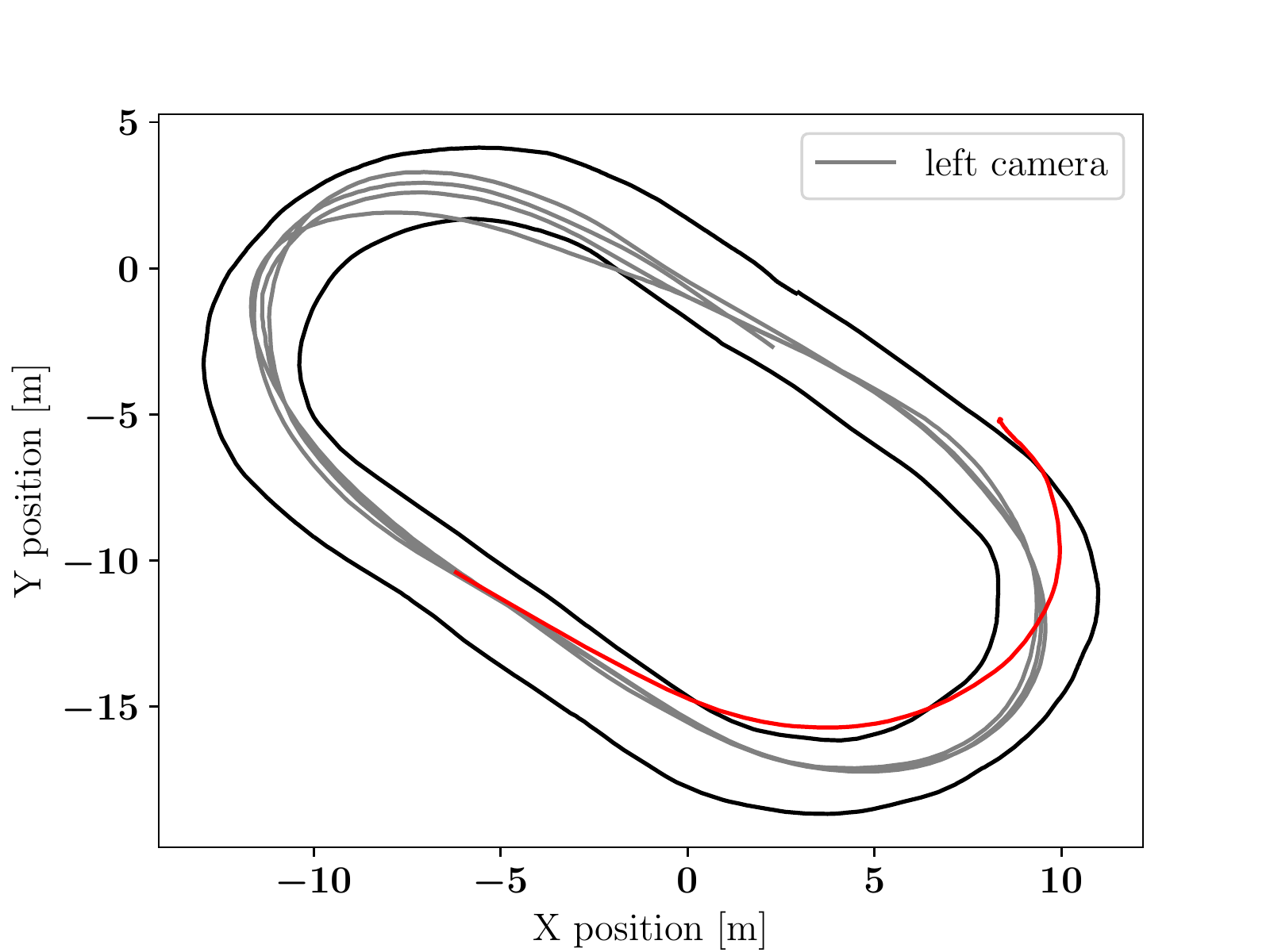}}
    \hfil
    \subfloat[Right camera \ac{NN} \label{fig:img_right}]{\includegraphics[width=0.24\textwidth]{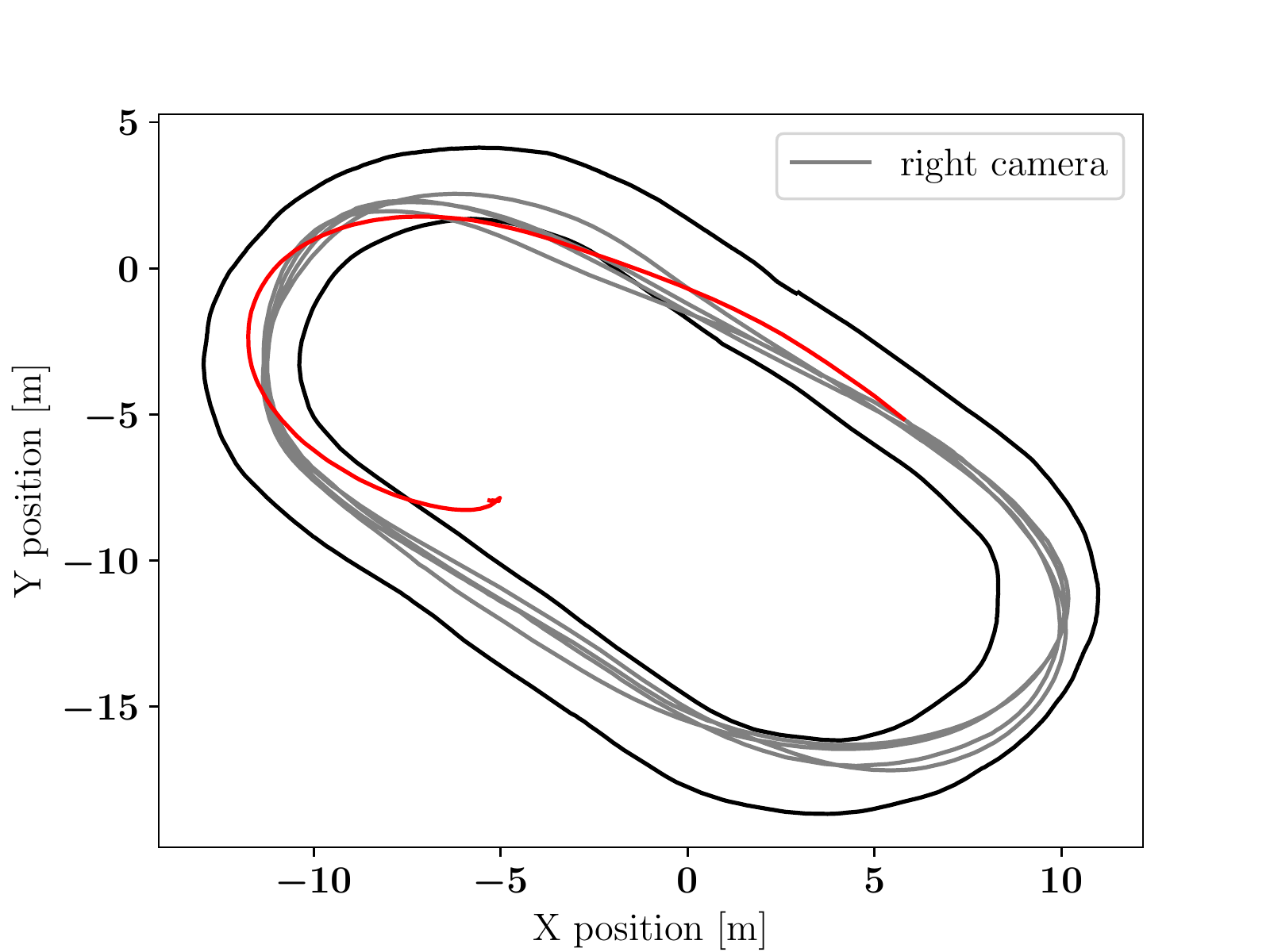}}
    \subfloat[GPS \ac{NN}\label{fig:img_gps}]{\includegraphics[width=0.24\textwidth]{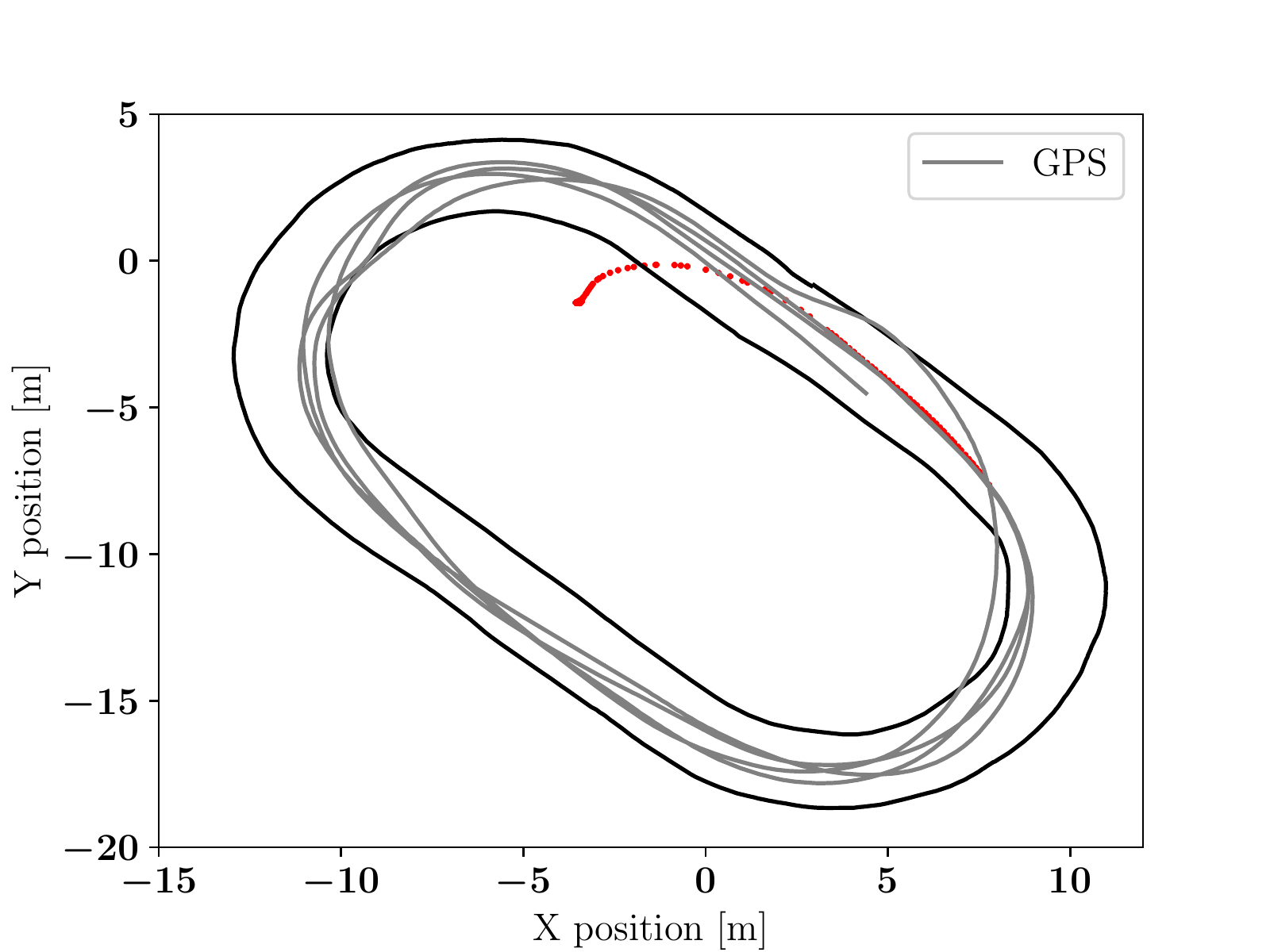}}
    \caption{Trajectory plots of imitation learning of autonomous driving with injected noise. The red trajectories are after the noise injected to each model's input after 4 laps of autonomous driving. We ran 10 experiments for each and all end-to-end learners immediately failed the task.}\label{fig:failures}
\end{figure}

Next, each learner in the ensemble was tested individually on the vehicle with artificial noise injected. After 4 laps of normal operation, noise was added to the corresponding sensor and crashes occurred immediately, as shown in \cref{fig:img_left}, \cref{fig:img_right} and \cref{fig:img_gps}. The test was repeated 10 times for each learner and crashes followed promptly after noise injection every time.

Following this experiment, the Ensemble \acl{BNN}s algorithm was tested without noise injection. The vehicle achieved similar performance to the expert, as seen in \cref{fig:mpcddp}, and was able to run at a high speed with no crashes.

\begin{figure*}[h]
    \centering
    \subfloat[1-4th, 7-8th, 11-12th, and 15-17th laps of trajectory plots of Ensemble Bayesian decision making without any artificial noise injected. \label{fig:no noise}]{\includegraphics[width=0.37\textwidth]{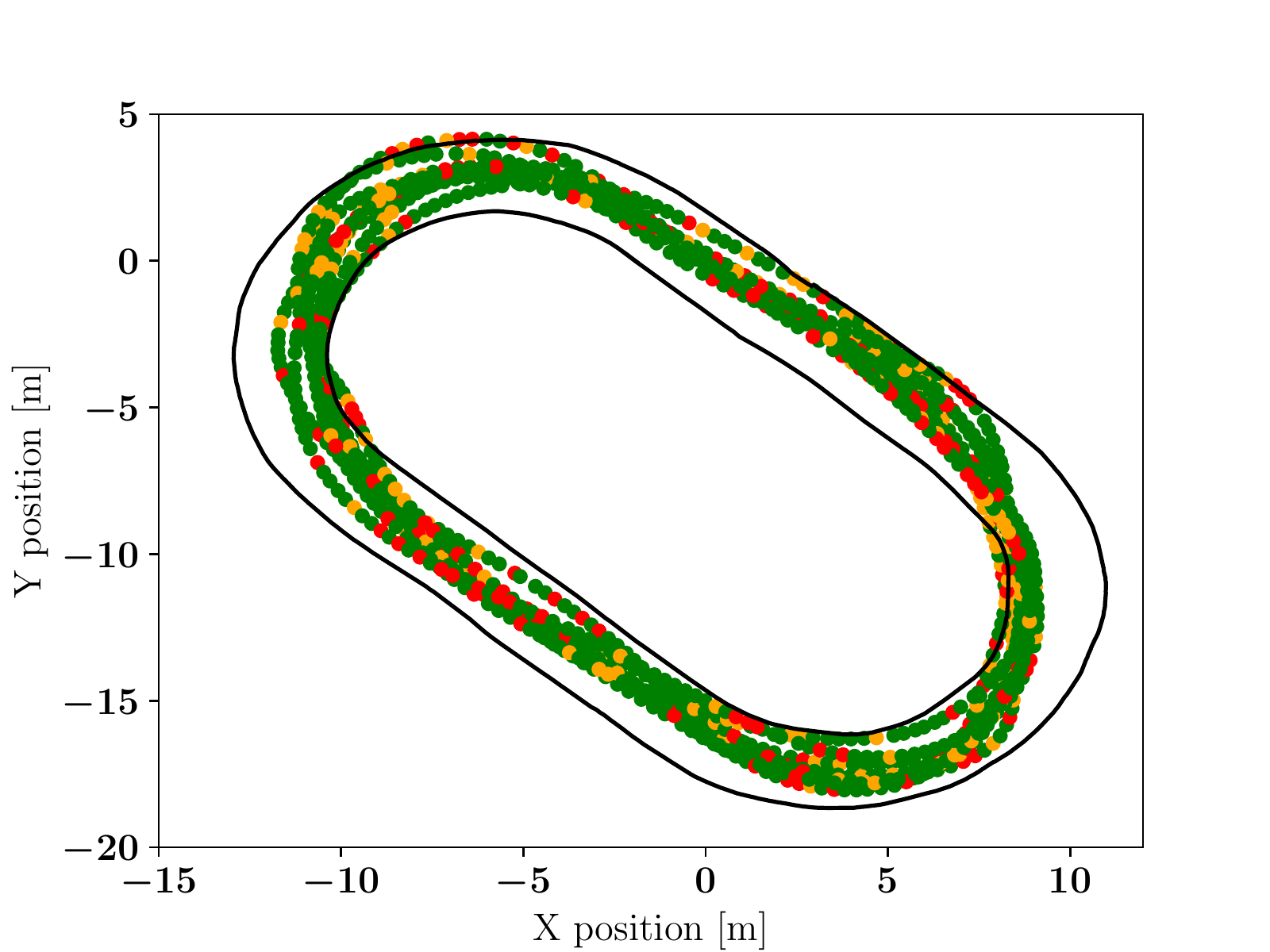}\includegraphics[width=0.105\textwidth]{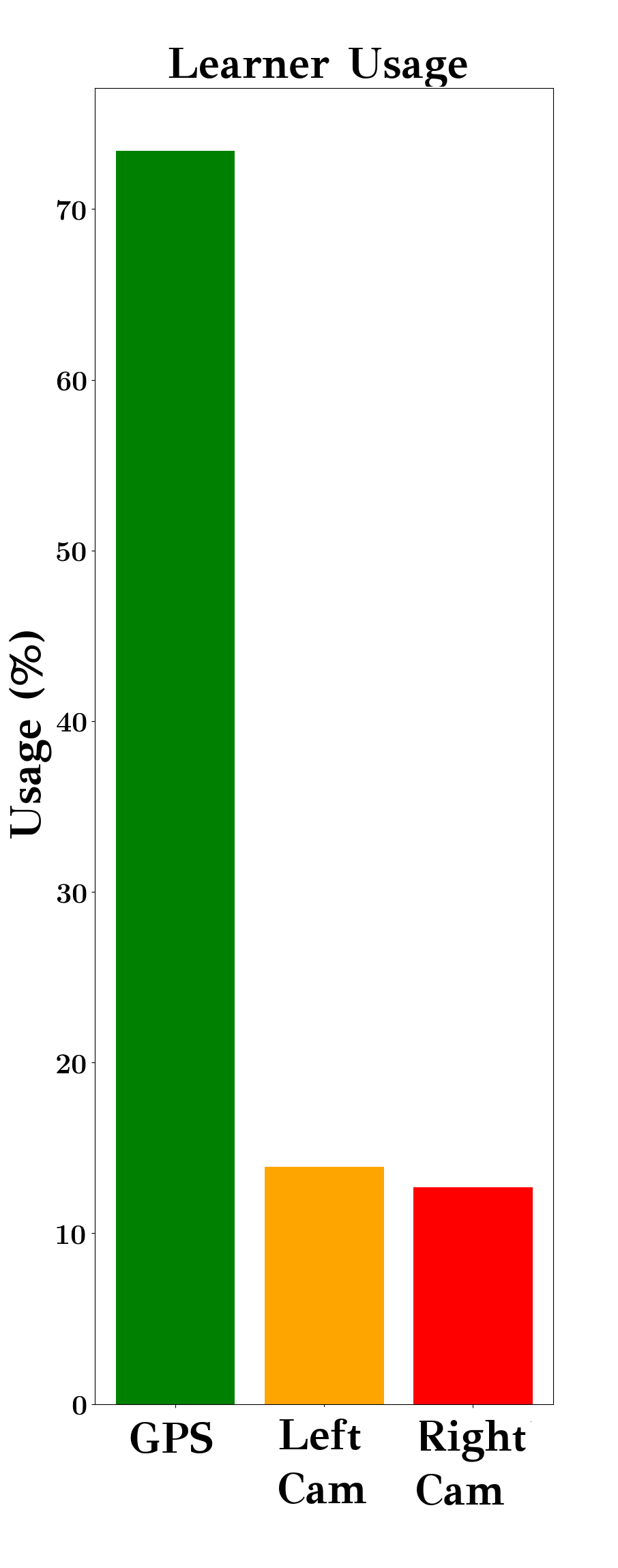}}
    \quad
    \subfloat[5-6th laps with frequent noise injected in position x and y data in the GPS signal. \label{fig:GPS noise}]{\includegraphics[width=0.37\textwidth]{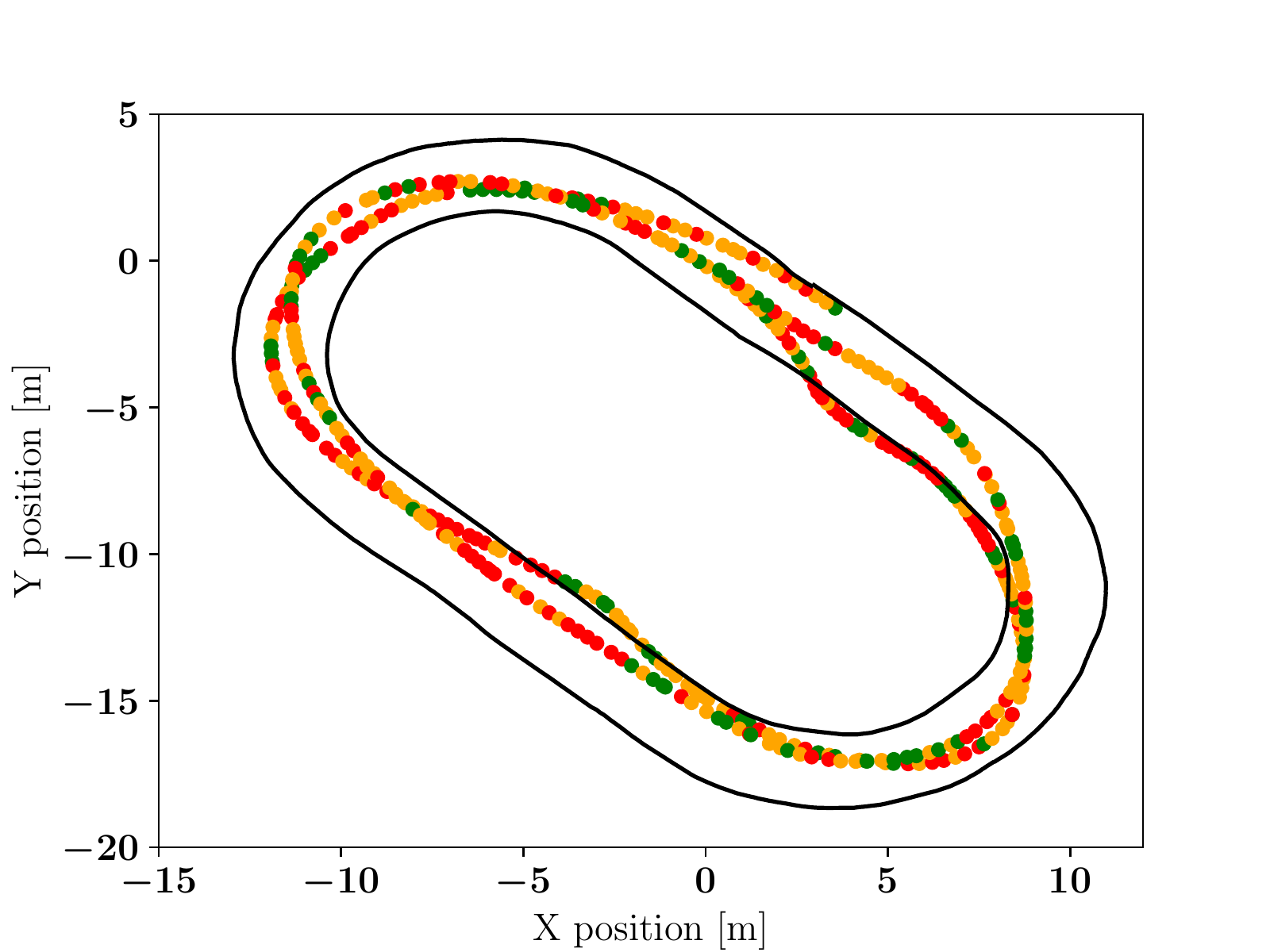} \includegraphics[width=0.105\textwidth]{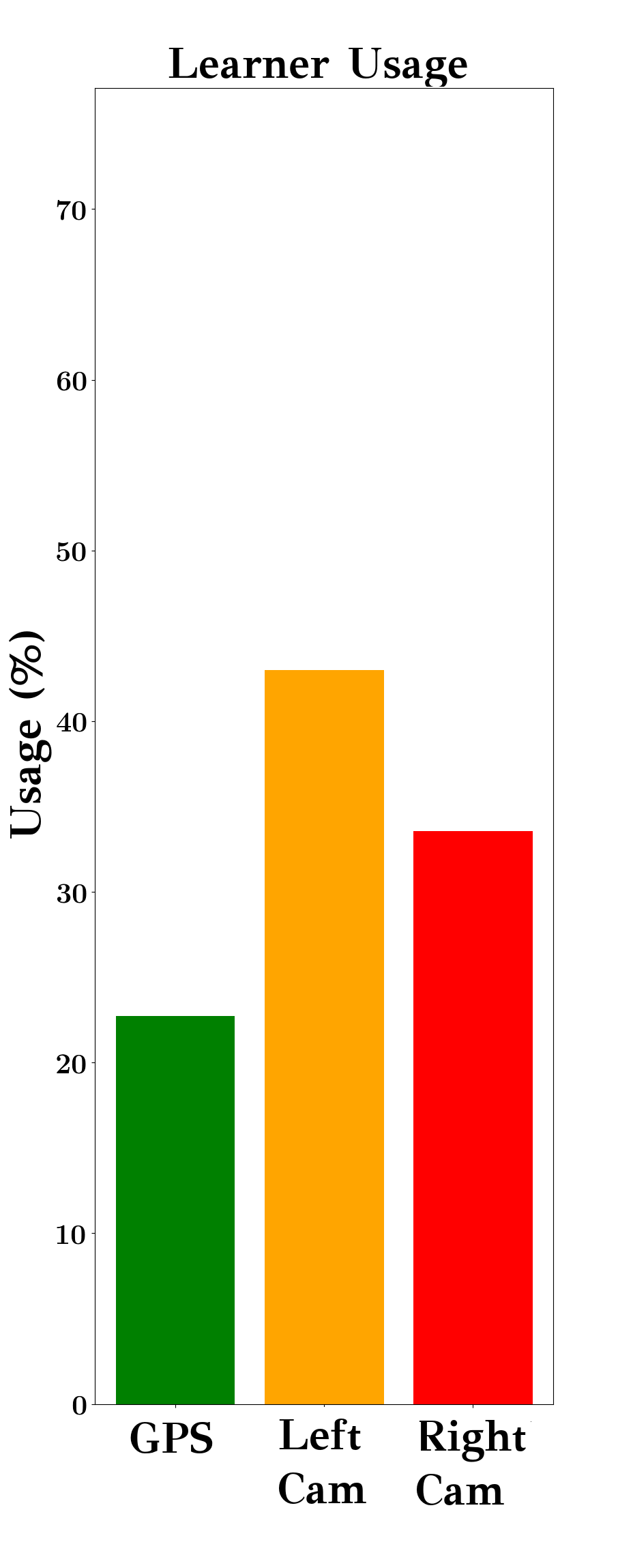}}
    \hfil
    \subfloat[9-10th laps with frequent noise injected in the left camera.\label{fig:left cam noise}]{\includegraphics[width=0.37\textwidth]{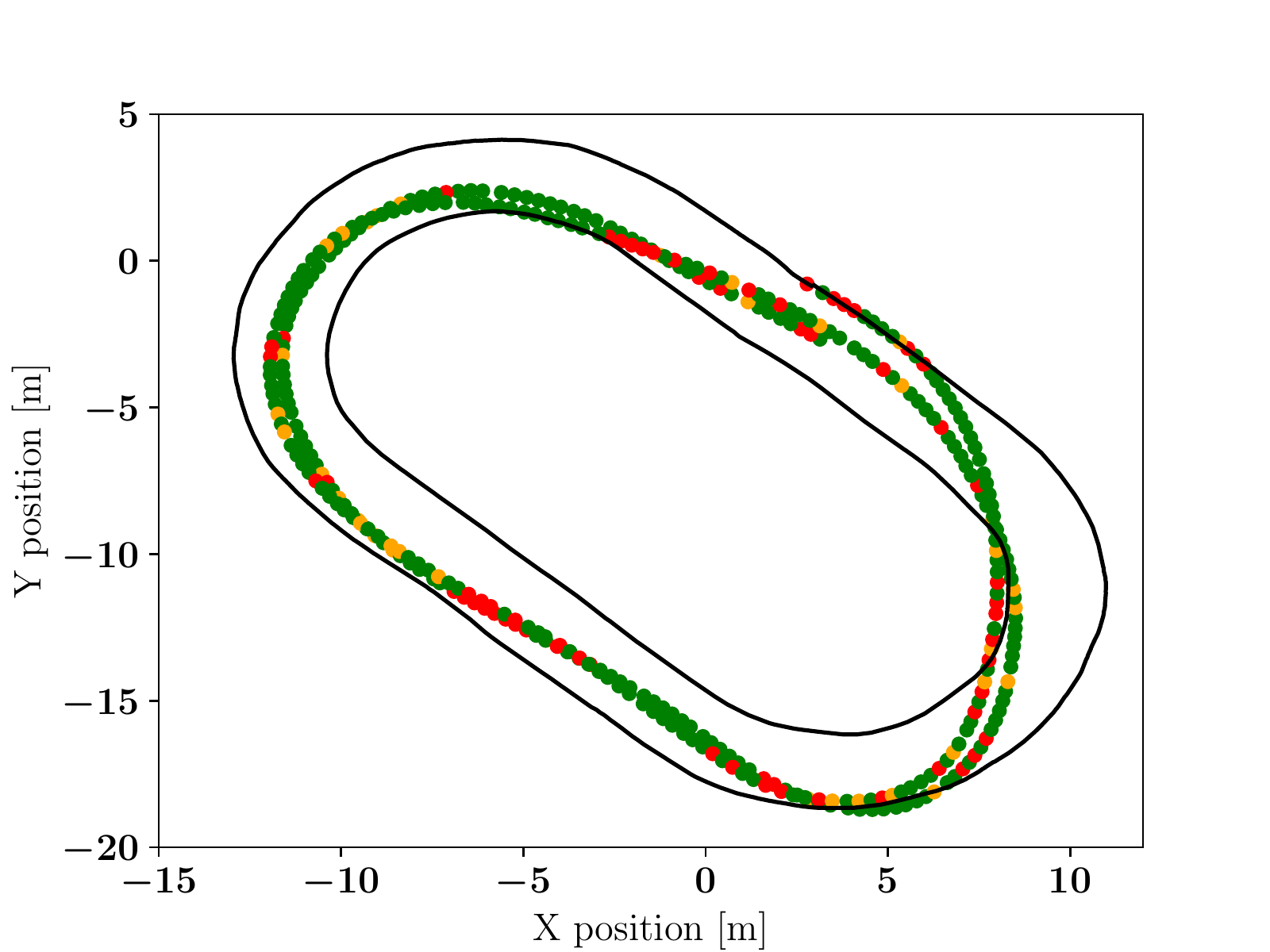} \includegraphics[width=0.105\textwidth]{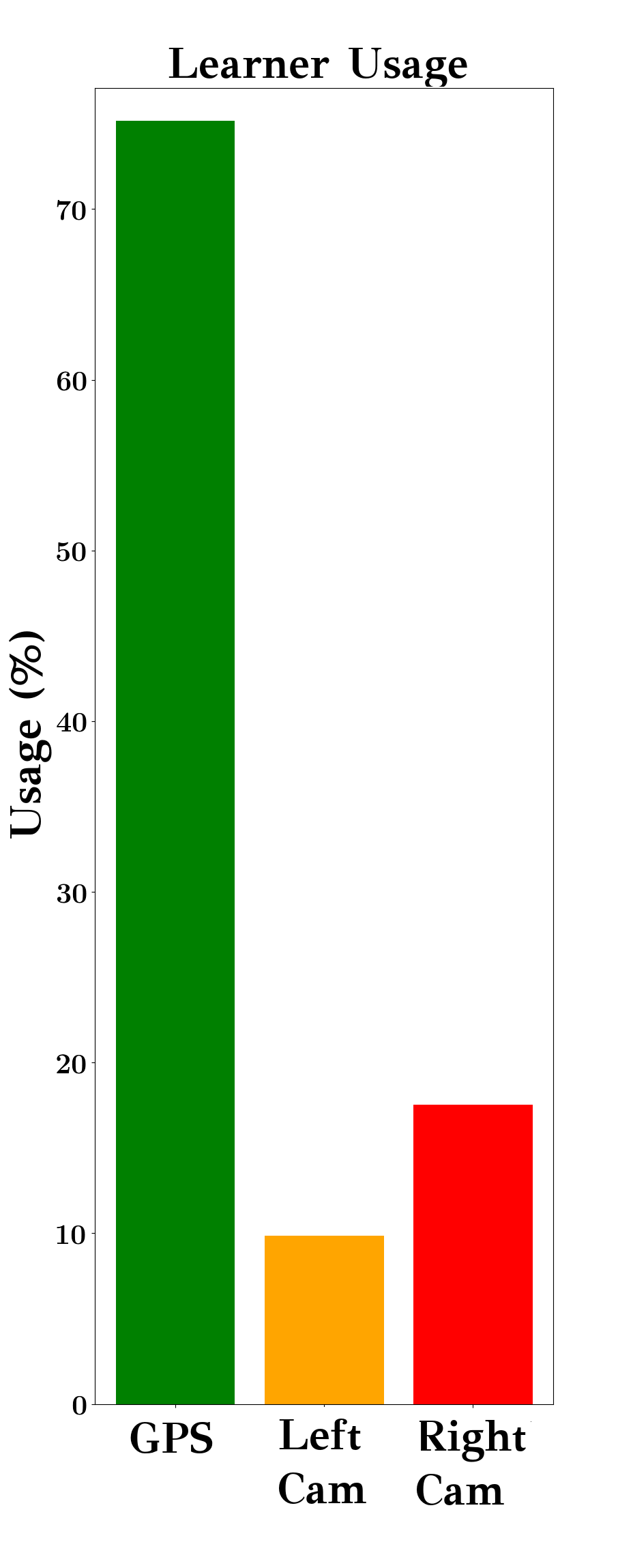}}
    \quad
    \subfloat[13-14th laps with frequent noise injected in both cameras.\label{fig:right cam noise}]{\includegraphics[width=0.37\textwidth]{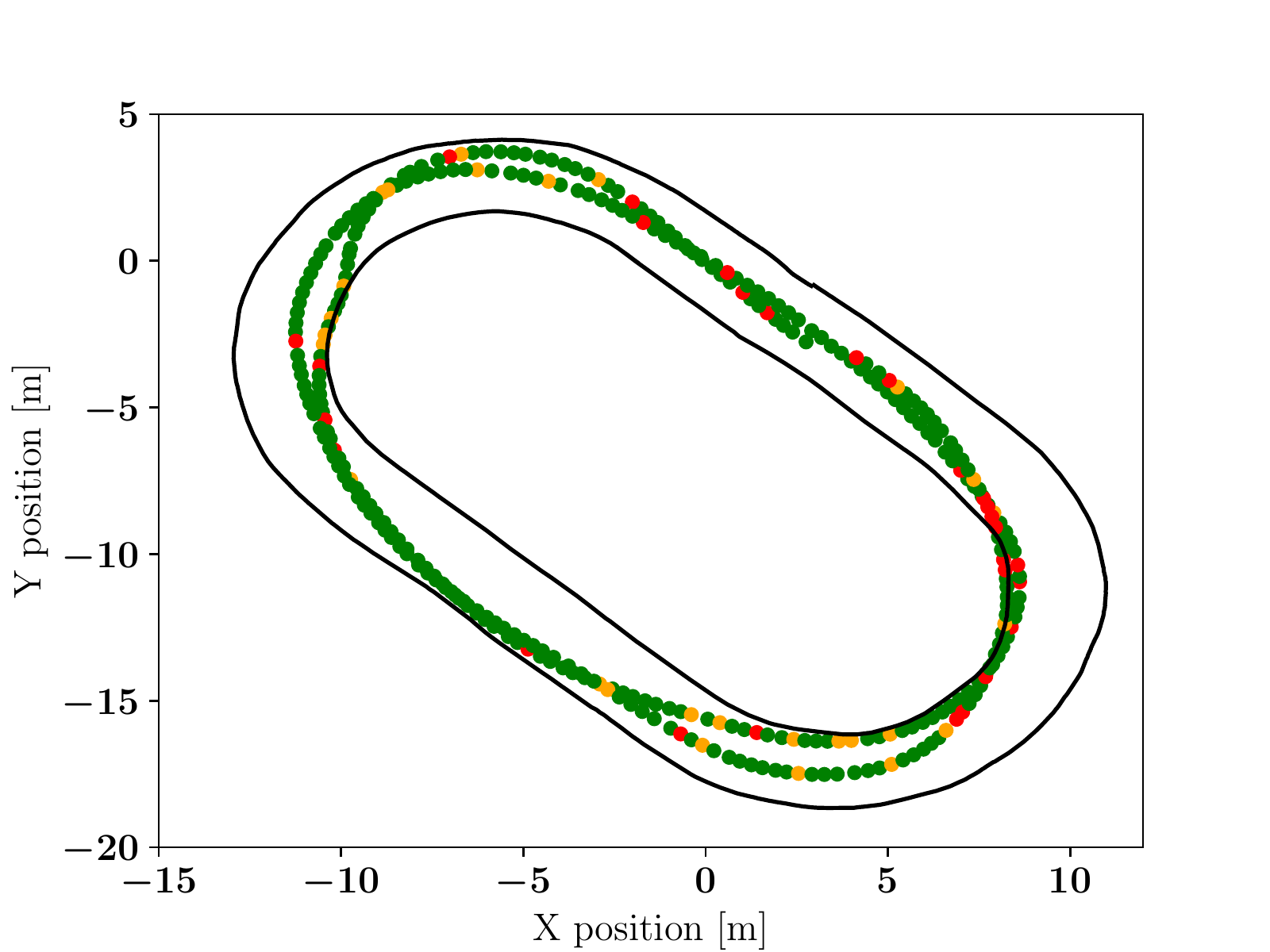} \includegraphics[width=0.105\textwidth]{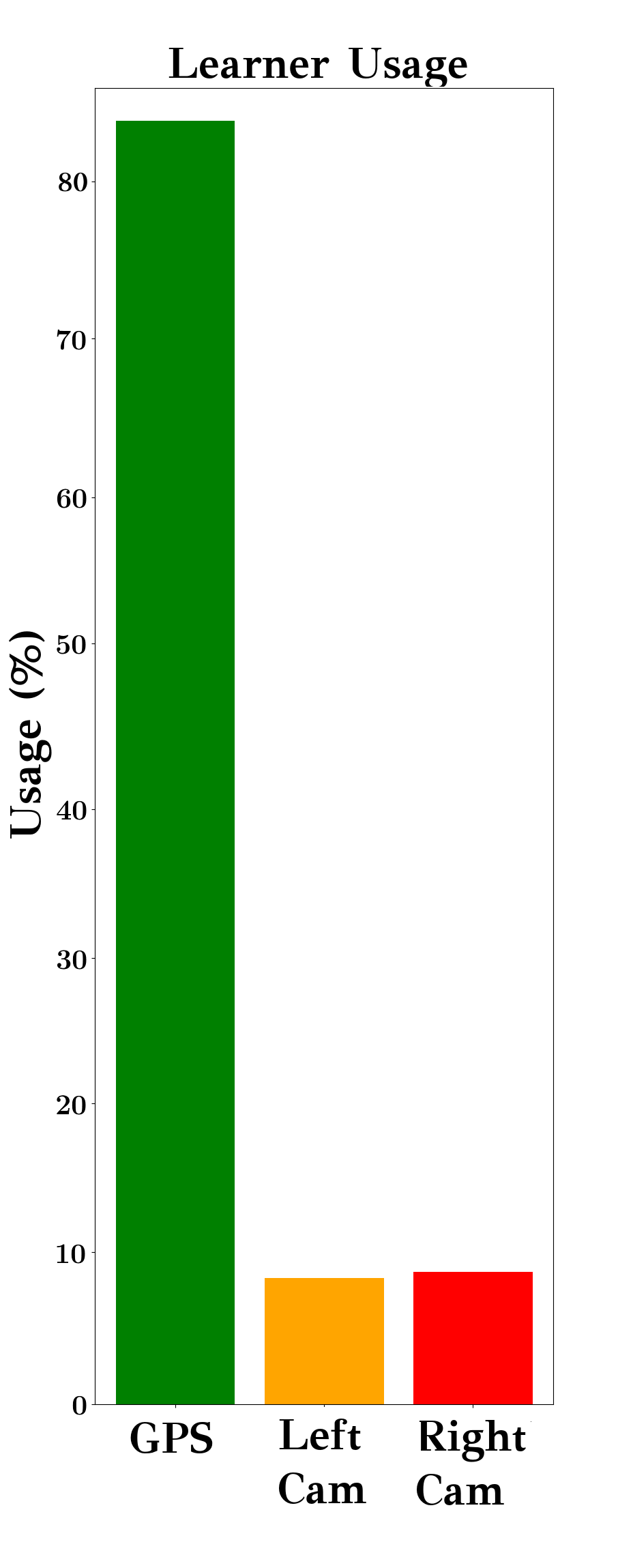}}
    \caption{Each lap is plotted with the colors of the learner whose action has been chosen: \textcolor{OliveGreen}{GPS \ac{NN}}, \textcolor{Orange}{Left camera \ac{NN}} and \textcolor{red}{Right camera \ac{NN}}. The algorithm was tested on the track 3 times for a total of 51 laps without failure of the task. }
\end{figure*}

Finally, the Ensemble \ac{BNN} algorithm was tested with noise injection. The time horizon for testing was set to be 17 laps. The algorithm was tested on the track 3 times for a total of 51 laps. After 4 laps of normal operation (\cref{fig:no noise}), frequent noise was added to each sensor in the order of GPS, left camera, and right camera for 2 laps. Normal operation resumed for another 2 laps before the next noise injection.
As we can see from the normal operation case in \cref{fig:no noise}, GPS \ac{NN} was usually used for most of the time. This is because the structure of the GPS \ac{NN} and the data (7 states) used for it were not complex as the structure of the Dropout VGG 16 network \cite{lee2019earlyfailure} and the data (RGB image) used for it. With a simpler structure and data, it is reasonable to have a smaller variance from the probabilistic network after training. Moreover, without any injection of artificial noise, GPS data at test time does not change much compared to the training data whereas the image from the camera slightly changes due to the change of the lighting conditions and the environment around the vehicle.
\cref{fig:GPS noise} shows that when artificial GPS noise was added, the algorithm opted to use camera inputs for navigation as a result of orders of magnitude increase in prediction uncertainty from the fully connected GPS \ac{NN}. For both normal case and GPS-noise injected case, we observe that the left camera \ac{NN} was used more often than the right camera \ac{NN}. We believe this behavior is task-specific, as the vehicle run the oval track in counterclockwise for both data-collecting and testing. Since the left camera is able to see the left track boundary more often than the right camera does, the left camera \ac{NN} is more confident about its prediction, resulting in smaller variance.
\cref{fig:left cam noise} demonstrated a decrease in usage of the left camera input, since image noise caused uncertainty from the corresponding network to double. Similar results can be found in \cref{fig:right cam noise} when image noise was added to both left and right cameras. Compared to the cases where we did not inject any noise (\cref{fig:no noise}) or noise was injected in a single camera (\cref{fig:left cam noise}), we can see the decreased usage of both cameras. For all cases of noise injection scenarios, the noise was injected frequently, but not always, so the noise-injected learner could be used intermittently when the noise did not exist. The usage of each learner with sensor noise injection is listed in \cref{tab:usage}. In all cases, the usage of the sensor(s) was decreased when the artificial noise was injected to the sensor(s). Even with large noise, which causes the immediate failure of the task for an individual \ac{BNN}, all laps were completed without any failure.
The complete trial can be seen in the video online\footnote{\href{https://youtu.be/poRbH__kB2M}{https://youtu.be/poRbH$\_\_$kB2M}}.

\begin{table}[h]
\caption{Learner Usage On Each Lap}
\label{tab:usage}
\begin{tabular}{llllllll}
Laps                    & 1-4, 7-8, 11-12    & 5-6    & 9-10     & 13-14  \\
\hline
Noise injected in       &  -                        & GPS    & Left Cam & Both Cams \\
GPS \ac{NN}($\%$)       & 73.4                      & \textbf{22.7}   & 72.5     & 83.6 \\
Left Cam \ac{NN}($\%$)  & 13.9                      & 42.7   & \textbf{10.0}     & \textbf{8.1} \\
Right Cam \ac{NN}($\%$) & 12.7                      & 34.6   & 17.5     & \textbf{8.3}
\end{tabular}
\end{table}

\section{Conclusions}\label{sec:conclusion}
In this paper, we introduced an Ensemble \acl{BNN} structure for system redundancy in the decision making of safety-critical systems. Our algorithm was implemented on an autonomous driving task using end-to-end \acl{IL}. Prediction uncertainty capturing both model imperfection and data insufficiency of each \ac{BNN} within the ensemble was used to switch between the different policy outputs. Experimental results verified the robustness of our proposed method against compromised sensor inputs. Our method can play an important role in any kind of autonomous systems using multiple sensors, especially in dealing with safety-critical tasks.

For future works on Ensemble Bayesian decision making, we will further investigate the switching mechanism in the ensemble to ensure safe and stable operation during switching. Furthermore, we will explore smooth Bayesian mixing models as an alternative to our current switching mechanism. Finally, we would like to also extend this Ensemble Bayesian approach to robust filtering and state estimation problems, where we use multiple sensors or networks.

\bibliographystyle{IEEEtran}
\bibliography{ensemble}

\onecolumn
\section*{Citations}
\hspace{-0.3cm}Plain Text:
\\ \\
K. Lee, Z. Wang, B. Vlahov, H. Brar, and E. A. Theodorou, ``Ensemble Bayesian Decision Making with Redundant Deep Perceptual Control Policies,'' The 18th IEEE International Conference on Machine Learning and Applications, 2019.
\\ \\
BibTeX:
\\ \\
@ARTICLE$\{$lee2019ensemble,
\\author=$\{$Keuntaek $\{$Lee$\}$ and Ziyi $\{$Wang$\}$ and Bogdan $\{$Vlahov$\}$ and Harleen $\{$Brar$\}$ and Evangelos A. $\{$Theodorou$\}$$\}$,
\\journal=$\{$The 18th IEEE International Conference on Machine Learning and Applications$\}$,
\\title=$\{$$\{$Ensemble Bayesian Decision Making with Redundant Deep Perceptual Control Policies$\}$$\}$,
\\year=$\{$2019$\}$
\\$\}$

\end{document}